\documentclass[conference]{IEEEtran}
\usepackage{cite}
\usepackage{amsmath,amssymb,amsfonts}
\usepackage{algorithmic}
\usepackage{graphicx}
\usepackage{textcomp}
\usepackage{xcolor}
\def\BibTeX{{\rm B\kern-.05em{\sc i\kern-.025em b}\kern-.08em
    T\kern-.1667em\lower.7ex\hbox{E}\kern-.125emX}}
\usepackage{graphicx}

\usepackage{pgfplots}
\usepackage{pgfplotstable}
\usepackage{booktabs, colortbl, array}
\usepackage{enumitem}
\usepackage{rotating}
\usepackage{balance}
\usepackage{bbm}
\usepackage{amssymb}
\usepackage{amsmath}
\usepackage{amsthm}

\usepackage{url}

\begin{document}
\title{DM algorithms in healthindustry}
%
%
\author{\IEEEauthorblockN{Li Wang}
\IEEEauthorblockA{\textit{Wufong University, China}
}}
\maketitle              
\begin{abstract}
This survey reviews several approaches of data mining (DM) in healthindustry from many research groups world wide. The focus is on modern multi-core processors built into today's commodity computers, which are typically found at university institutes both as small server and workstation computers. So they are deliberately not high-performance computers. Modern multi-core processors consist of several (2 to over 100) computer cores, which work independently of each other according to the principle of ``multiple instruction multiple data'' (MIMD). They have a common main memory (shared memory). Each of these computer cores has several (2-16) arithmetic-logic units, which can simultaneously carry out the same arithmetic operation on several data in a vector-like manner (single instruction multiple data, SIMD). DM algorithms must use both types of parallelism (SIMD and MIMD), with access to the main memory (centralized component) being the main barrier to increased efficiency. This is important for DM in healthindustry applications like ECG, EEG, CT, SPECT, fMRI, DTI, ultrasound, microscopy, dermascopy, etc.
\end{abstract}

\section{Introduction}
Healthindustry applications like ECG, EEG, CT, SPECT, fMRI, DTI, ultrasound, microscopy, dermascopy, etc. pose high requirements on the runtime performance of data mining (DM).
Modern computer hardware supports the development of high-performance applications for data analysis on many different levels. The focus is on modern multi-core processors built into today's commodity computers, which are typically found at university institutes both as small server and workstation computers. So they are deliberately not high-performance computers. Modern multi-core processors consist of several (2 to over 100) computer cores, which work independently of each other according to the principle of ``multiple instruction multiple data'' (MIMD). They have a common main memory (shared memory). Each of these computer cores has several (2-16) arithmetic-logic units, which can simultaneously carry out the same arithmetic operation on several data in a vector-like manner (single instruction multiple data, SIMD). DM algorithms must use both types of parallelism (SIMD and MIMD), with access to the main memory (centralized component) being the main barrier to increased efficiency. In principle, fast caches allow this problem to be solved. However, high-performance algorithms must be designed in such a way that these caches can work effectively (cache-conscious or cache-oblivious).

For standard tasks, e.g. in linear algebra, optimization or training of neural networks, high performance algorithms with SIMD and MIMD parallelism (partly with efficient cache usage) are already widely used. More specialized methods, such as for cluster analysis, require a specific new development of the algorithms, as there have published in numerous articles on the cluster methods k-Means, EM and DBSCAN and other algorithms for GPU and multi-core architectures. This is mainly about the development of general algorithmic principles that can be applied to many problems, e.g. proposals for cache-oblivious loops. The idea of cache-oblivious loops is to replace nested loops (which play a central role in many application algorithms) in such a way that they (1) can be distributed to parallel units in a locality preserving manner and (2) the accesses themselves are designed in such a way that a high degree of locality and thus efficient cache usage is achieved. This is achieved by using a special variant of the Hilbert curve (a space-filling curve called FUR-Hilbert). Compared to the basic version of the algorithms (without SIMD and MIMD parallelism), it is often possible to have acceleration factors of two or more orders of magnitude (i.e. over 100).

The further development of these parallelization techniques for data analysis algorithms, as well as their exemplary use in the lead applications of healthcare and life sciences is the overall goal of this research area.

An interesting goal would be to further develop the basic algorithmic paradigm of defining high-performance algorithms for data analysis using the concept of cache-oblivious loops with the following goals:
\begin{itemize}
\item The performance shall be further increased compared to the previous approach, e.g. by dynamic load balancing.
\item The usability for various new application algorithms shall be improved, e.g. by developing specialized algorithms to detect data dependencies and to develop convergence criteria for such algorithms
\item The concept of cache-oblivious loops is to be developed for further architectures, e.g. GPU or distributed systems (e.g. grid and cloud environments).
\end{itemize}

\section{Cache-oblivious loops with dynamic load balancing}
A SIMD and MIMD-based parallelization approach to design loop pairs in such a way that they flexibly support caches of any size divides the total workload of an algorithm by means of so-called space-filling curves into packets that are assigned to the individual cores and make efficient use of the cache memory there. Although the workload can initially be optimally distributed among the computer cores in this way in a very simple manner, it is not yet possible to dynamically rebalance the load if, for example, the cores are subject to an uneven basic load due to external influences. In this work package, procedures for dynamic load redistribution must be designed in such a way that they do not collide with the overall goal of efficient cache usage.

\section{Modelling of loop structures}
The current approach is limited to traversing two nested loops with fixed loop boundaries in a way that uses caches of any size and also allows MIMD parallelism. A newly developed variant of the Hilbert curve was used for this purpose. In this work package this concept shall be extended to any number of loops and dependencies of the loops among each other shall be modelable. This should make it possible to efficiently process e.g. triangular or band matrices as well as higher order tensors. For the processing of three or more nested loops, the mathematical extensions of the Hilbert curve and other space-filling curves into higher-dimensional spaces basically already exist. However, these methods still have to be extended and adapted in such a way that the calculation can be performed highly efficiently (in constant time per loop pass), which were achieved in the two-dimensional case with the help of different concepts. The extension of these concepts for spaces of arbitrary dimensionality seems to us as a research task demanding but also feasible. For the modelling of dependencies it is possible to define suitable data structures that are oriented towards and support the hierarchical-recursive nature of the space-filling curves.

\section{Modelling of monotony requirements}
This approach, based on a special variant of the Hilbert curve, offers the advantage that in most applications the effect on cache location is strongest compared to comparable space-filling curves. However, some application algorithms require a sequence of loop passes that fulfills certain monotony properties, i.e. certain loop indices are processed before certain other loop indices. The idea is to combine existing space-filling curves such as Hilbert or the Z-order in such a way that even partial monotony requirements can be represented in individual dimensions. In this work package, the concept of cache-oblivious loops will be extended to specify monotonicity properties and automatically apply suitable space-filling curves.

\section{GPU and Distributed Systems}
The problem that algorithms work distributed on data in such a way that each processing unit reaches the highest possible access locality does not only arise in multi-core CPU systems, but also in other distributed and parallel architectures. The setting is very similar for graphic processing units (GPU), but the basic problem is similar for distributed systems without shared physical memory (grids, cloud computing). In this work package the extension to such architectures will be investigated. While the basic problem of locality preservation remains the same, in alternative hardware scenarios the considerations about the trade-off between different cost factors of the algorithms, i.e. especially the trade-off between the transmission costs of the data over the network or internal bus connections and the computing time (which also includes the management of space-filling curves and other techniques for locality preservation) can vary in detail.

\section{Derivation of criteria for applicability}
Not all data analysis algorithms are equally suitable for parallelization using cache-oblivious loops, even if they follow the basic pattern of nested loops. Therefore, in this work package a set of criteria is developed to find data dependencies and monotony requirements in the application algorithms manually or (semi-)automatically. In addition, criteria are to be developed to determine when algorithms cannot be transformed equivalently in a provable way, but after a transformation, algorithms result which are also convergent and possibly reach a local optimum deviating from the original algorithm. From a scientific point of view, the latter is even the greater challenge, because automatic techniques for the recognition of data dependencies have already been proposed and are very successfully and widely used in other sub-areas of computer science (e.g. compiler construction). On the other hand, techniques for transformation into merely result-equivalent algorithms are much less known.

\section{Transformation of important application algorithms}
Vrious algorithms are to be transformed exemplarily and to support the development of criteria above in order to be applicable in a SIMD and MIMD parallel architecture with the help of cache-oblivious loops. Very general building blocks such as different matrix decompositions (LU, QR, Jacobi) support dimension transforming or reducing methods such as PCA, Kernel-PCA, ICA or t-SNE, as they are used in the lead application of medical image processing (different modalities of magnetic resonance imaging). In addition, algorithms from network planning techniques for planning in traffic networks or for industrial production planning are also being considered, for example.
\section{Related Work}

There are various approaches for \textbf{different data types}. The data can be of any type, as long as a distance function exists. Fixed-length text data often uses Hamming distance \cite{DBLP:journals/tjs/HoOK18} and the similarity between variable length text is often measured using the edit distance  \cite{DBLP:journals/pvldb/XiaoWL08}. A common measure for set data is the Jaccard distance  \cite{DBLP:conf/sigmod/DengT018,DBLP:journals/tods/XiaoWLYW11}, whereas the similarity of documents is processed with cosine-like similarity measures \cite{DBLP:series/synthesis/2013Augsten,DBLP:conf/icde/ShangLLF17}.

\textbf{Approximate} nearest neighbor search techniques can also be applied to the similarity join problem, however without guarantees on completeness and exactness of the result. There may be false positives as well as false negatives. Recently an approach \cite{DBLP:journals/tkde/YuNLWY17} to Locality Sensitive Hashing (LSH) is used on a representative point sample, to reduce the number of lookup operations. LSH is of interest in theoretical foundational work, where a recursive and cache-oblivous LSH approach \cite{DBLP:journals/algorithmica/PaghPSS17} was proposed. The topic of approximate solutions for the similarity join is also an emerging field in deep learning \cite{DBLP:journals/corr/abs-1803-04765}. There are approximative approaches which target low dimensional cases (spatial joins in 2--3 dimensions \cite{DBLP:conf/icde/BryanEF08}) or higher (10--20) dimensional cases  \cite{DBLP:conf/focs/AndoniI06}. Very high-dimensional cases, with dimensions of $128$ and above have been targeted with Symbolic Aggregate approXimation (SAX) techniques \cite{DBLP:journals/concurrency/MaJZ17}) to generate approximate candidates. SAX techniques rely on several indirect parameters like PAA size or the iSAX alphabet size.

There are preconstructed indexing techniques, which are based on \textbf{space-filling curves} and applied to the similarity join problem. Specifically, where the data is sorted efficiently with respect to one or more Z-order curves \cite{DBLP:conf/kdd/DittrichS01, DBLP:journals/tkde/KoudasS00, DBLP:conf/icde/LiebermanSS08} in order to test the intersection of the hypercubes in the datastructures. Others propose space-filling curves, to reduce the storage cost for the index \cite{DBLP:journals/tkde/ChenGLJC17}.
LESS \cite{DBLP:conf/icde/LiebermanSS08} targets GPUs and not multi-core environments. ZC and MSJ \cite{DBLP:journals/tkde/KoudasS00} as well as the SPB-tree index \cite{DBLP:journals/tkde/ChenGLJC17}, although simple, they require space transformations and preprocessing, which make them hard to parallelize.

\textbf{EGO family} of $\epsilon$-join algorithms. The EGO-join algorithm is the first algorithm in this family introduced by B\"{o}hm et al. in  \cite{epsilongridorder}. The Epsilon Grid Order (EGO) was introduced as a strict order (i.e. an order which is irreflexive, asymmetric and transitive). It was shown that all join partners of some point $\mathbf x$ lie within an $\epsilon$-interval, of the Epsilon Grid Order. Algorithms of the EGO family exploit this knowledge for the join operation. The EGO-join has been re-implemented as a recursive variant with additional heuristics, to quickly decide whether two sequences are non-join-able \cite{DBLP:conf/dasfaa/KalashnikovP03}. Further improvements proposed two new members of this family, the EGO$^{*}$ \cite{DBLP:journals/is/KalashnikovP07} algorithm and its extended version called Super-EGO \cite{DBLP:journals/vldb/Kalashnikov13} target multi-core environments using a multi-process/multi-thread programming model. Super-EGO proposes a dimensional reordering \cite{DBLP:journals/vldb/Kalashnikov13}. In the experiments Super-EGO encounters some difficulties with uniformly distributed data, particularly when the number of data objects exceeds millions of points or the dimensionality is above $32$.


If the similarity join runs multiple times on the same instances of the data, one might consider \textbf{index-based approaches} \cite{DBLP:conf/icde/BohmK01, DBLP:journals/jda/ParedesR09, DBLP:journals/tkde/ChenGLJC17}, such as R-tree \cite{DBLP:conf/sigmod/BrinkhoffKS93} or \textit{M}-tree \cite{DBLP:conf/vldb/CiacciaPZ97}. Index-based approaches have the potential to reduce the execution time, since the index stores pre-computed information that significantly reduces query execution time. This pre-computational step could be costly, especially in the case of List of Twin Clusters (LTC)
\cite{DBLP:journals/jda/ParedesR09}, where the algorithm needs to build joint or combined indices for every pair of points in the dataset. The D-Index \cite{DBLP:journals/mta/DohnalGSZ03} and its extensions (i.e. eD-Index \cite{DBLP:conf/dexa/DohnalGZ03} or i-Sim index \cite{DBLP:conf/sisap/PearsonS14}) build a hierachical structure of index levels, where each level is organized into separable buckets and an exclusion set. The most important drawback of D-Index, eD-Index and i-Sim is that they may require rebuilding the index structure for different $\epsilon$.

\textbf{Data partitioning across multiple machines} is not the main focus of this paper where we assume that the data fits into main memory. The case of relational join algorithms has been studied extensively in the past \cite{DBLP:conf/sigmod/SchneiderD89, DBLP:conf/kdd/WangMP13, DBLP:journals/pvldb/FierABLF18}.
The similarity join has been successfully applied in the distributed environment with different MapReduce variants \cite{DBLP:conf/waim/LiWU16, DBLP:conf/sigmod/McCauley018, DBLP:journals/pvldb/FierABLF18}. Another distributed version is proposed in \cite{DBLP:conf/sigmod/ZhaoRDW16}. There, a multi-node solution with load-balancing is used, that does not require re-partitioning on the input data. This variant focuses on minimization of data transfer, network congestion and load-balancing across multiple nodes.

The similarity join has been already implemented for \textbf{Graphics Processing Units (GPUs)}.
In \cite{DBLP:conf/btw/BohmNPZ09} the authors use a directory structure to generate candidate points. On datasets with 8 million points, the proposed GPU algorithm is faster than its CPU variant, when the
 $\epsilon$-region has at least 1 or 2 average neighbors.
LSS \cite{DBLP:conf/icde/LiebermanSS08} is another similarity join variant for the GPU, which is suited for high dimensional data. Unfortunetly both \cite{DBLP:conf/icde/LiebermanSS08} and \cite{DBLP:conf/btw/BohmNPZ09}  are targeted to NVIDIA GPUs and have been optimized for an older version of CUDA.

\subsection{Cache-oblivious Algorithms}
\noindent Cache-oblivious algorithms \cite{DBLP:conf/focs/FrigoLPR99} have attracted considerable attention as they are portable to almost all environments and architectures. Algorithms and data structures for basic tasks like sorting, searching, or query processing \cite{DBLP:conf/sigmod/HeLLY07} and for specialized tasks like ray reordering \cite{DBLP:journals/tog/MoonBKCKBNY10} or homology search in bioinformatics \cite{DBLP:journals/bmcbi/FerreiraRR14} have been proposed. Two important algorithmic concepts of cache-oblivious algorithms are localized memory access and divide-and-conquer. The Hilbert curve integrates both ideas. The Hilbert curve defines a 1D ordering of the points of an 2-dimensional space such that each point is visited once. Bader et al. proposed to use the Peano curve for matrix multiplication and LU-decomposition \cite{DBLP:conf/para/BaderM06, DBLP:conf/europar/Bader08}. The algorithms process input matrices in a block-wise and recursive fashion where the Peano curve guides the processing order and thus the memory access pattern. In \cite{loopsjournal}, cache-oblivious loops have been applied to K-means clustering and matrix multiplication. 

\subsection{Optimized Techniques for Specific Tasks or Hardware}
\noindent The library BLAS (Basic Linear Algebra Subprograms) \cite{DBLP:journals/toms/DongarraCHD90} provides basic linear algebra operations together with programming interfaces to C and Fortran. BLAS is highly hardware optimized: specific implementations for various infrastructures are available, e.g. ACML for AMD Opteron processors or CUBLAS for NVIDIA GPUs. The Math Kernel Library (MKL) contains highly vectorized math processing routines for Intel processors. These implementations are very hardware-specific and mostly vendor-optimized. Moreover, they are designed to efficiently support specific linear algebra operations. Experiments demonstrate that the cache-oblivious approach reaches a performance better than BLAS on the task of the similarity join for points of dimensions in the range of $\{2,...,64\}$.

\section{Applications}
These techniques are investigated for DM algorithms in the following applications:
\begin{itemize}
\item Electrocardiography (ECG, 1D), 
\item Electroencephalography (EEG, 1D), 
\item Computer Tomography (CT, 3D, 4D), 
\item SPECT (4D), 
\item Structural Magnetic Resonance Imaging (MRI, 3D),
\item Functional Magnetic Resonance Imaging (fMRI, 4D), 
\item Diffusion Tensor Imaging (DTI, 6D), 
\item Ultrasound (US, 2D), 
\item Microscopy, Dermascopy (2D),
\item etc.
\end{itemize}

\section{Conclusion}
This survey reviews several approaches of DM from many research groups world wide. Modern computer hardware supports the development of high-performance applications for data analysis on many different levels. The focus is on modern multi-core processors built into today's commodity computers, which are typically found at university institutes both as small server and workstation computers. So they are deliberately not high-performance computers. Modern multi-core processors consist of several (2 to over 100) computer cores, which work independently of each other according to the principle of ``multiple instruction multiple data'' (MIMD). They have a common main memory (shared memory). Each of these computer cores has several (2-16) arithmetic-logic units, which can simultaneously carry out the same arithmetic operation on several data in a vector-like manner (single instruction multiple data, SIMD). DM algorithms must use both types of parallelism (SIMD and MIMD), with access to the main memory (centralized component) being the main barrier to increased efficiency. We investigate these performance issues in the context of healthindustry applications like ECG, EEG, CT, SPECT, fMRI, DTI, ultrasound, microscopy, dermascopy, etc.

\nocite{DBLP:conf/sigmod/BohmFP08,DBLP:conf/icdt/BerchtoldBKK01,DBLP:conf/icdm/BohmK02,10.1007/BFb0000120,DBLP:conf/icde/BohmOPY07,
DBLP:conf/cikm/BohmBBK00,DBLP:journals/jiis/BohmBKM00,DBLP:conf/adl/BohmBKS00,DBLP:conf/edbt/BohmK00,
DBLP:journals/sadm/AchtertBDKZ08,DBLP:journals/bioinformatics/BaumgartnerBBMWOLR04,DBLP:journals/jbi/BaumgartnerBB05,
DBLP:conf/icdt/BerchtoldBKK01,DBLP:conf/edbt/BohmP08,DBLP:conf/kdd/AchtertBKKZ06,DBLP:conf/cikm/BohmFOPW09,
DBLP:conf/ssdbm/BohmPS06,DBLP:conf/kdd/BohmHMP09,DBLP:conf/dawak/BerchtoldBKKX00,DBLP:conf/sdm/AchtertBDKZ08,
DBLP:journals/jdi/BaumgartnerGBF05,DBLP:journals/kais/MaiHFPB15,DBLP:journals/tlsdkcs/BohmNPWZ09,
DBLP:conf/dexa/BohmK03,DBLP:conf/icde/BohmGKPS07,DBLP:conf/miccai/DyrbaEWKPOMPBFFHKHKT12,DBLP:conf/kdd/PlantB11,
DBLP:journals/bioinformatics/PlantBTB06,DBLP:conf/icdm/ShaoPYB11,DBLP:conf/icdm/PlantWZ09,
DBLP:journals/tkdd/BohmFPP07,DBLP:conf/pakdd/BohmGOPPW10,DBLP:conf/btw/BohmNPZ09,DBLP:conf/icdm/MaiGP12,
DBLP:journals/kais/ShaoWYPB17,DBLP:conf/kdd/YeGPB16,DBLP:conf/kdd/FengHKBP12,DBLP:journals/envsoft/YangSSBP12,
DBLP:conf/icdm/GoeblHPB14,DBLP:conf/kdd/AltinigneliPB13,DBLP:conf/icdm/YeMHP16,DBLP:conf/kdd/Plant12,
DBLP:conf/kdd/Plant12,DBLP:conf/cikm/BohmBBK00,DBLP:journals/kais/BohmK04,DBLP:conf/icdm/BohmK02,
DBLP:conf/icdt/BerchtoldBKK01,loopsjournal,Bially1969SpacefillingCT,
Prusinkiewicz:1986:GAL:16564.16608,10.1007/BFb0000120,DBLP:conf/icdm/BohmK02,
DBLP:conf/ssdbm/AchtertBKKZ07,DBLP:conf/pkdd/AchtertBKKMZ06,SHAO20122756,DBLP:journals/tkde/ShaoHBYP13,DBLP:conf/icdm/BohmK02,
DBLP:conf/ssdbm/AchtertBKZ06,DBLP:conf/cikm/BohmBBK00,DBLP:journals/jiis/BohmBKM00}
\bibliographystyle{alpha}
\bibliography{bibliography, bib-new}

\newcommand{\etalchar}[1]{$^{#1}$}
\begin{thebibliography}{ABK{\etalchar{+}}06b}

\bibitem[AB13]{DBLP:series/synthesis/2013Augsten}
Nikolaus Augsten and Michael~H. B{\"{o}}hlen.
\newblock {\em Similarity Joins in Relational Database Systems}.
\newblock Synthesis Lectures on Data Management. Morgan {\&} Claypool
  Publishers, 2013.

\bibitem[ABD{\etalchar{+}}08a]{DBLP:journals/sadm/AchtertBDKZ08}
Elke Achtert, Christian B{\"{o}}hm, J{\"{o}}rn David, Peer Kr{\"{o}}ger, and
  Arthur Zimek.
\newblock Global correlation clustering based on the hough transform.
\newblock {\em Statistical Analysis and Data Mining}, 1(3):111--127, 2008.

\bibitem[ABD{\etalchar{+}}08b]{DBLP:conf/sdm/AchtertBDKZ08}
Elke Achtert, Christian B{\"{o}}hm, J{\"{o}}rn David, Peer Kr{\"{o}}ger, and
  Arthur Zimek.
\newblock Robust clustering in arbitrarily oriented subspaces.
\newblock In {\em Proceedings of the {SIAM} International Conference on Data
  Mining, {SDM} 2008, April 24-26, 2008, Atlanta, Georgia, {USA}}, pages
  763--774. {SIAM}, 2008.

\bibitem[ABK{\etalchar{+}}06a]{DBLP:conf/pkdd/AchtertBKKMZ06}
Elke Achtert, Christian B{\"{o}}hm, Hans{-}Peter Kriegel, Peer Kr{\"{o}}ger,
  Ina M{\"{u}}ller{-}Gorman, and Arthur Zimek.
\newblock Finding hierarchies of subspace clusters.
\newblock In {\em Knowledge Discovery in Databases: {PKDD} 2006, 10th European
  Conference on Principles and Practice of Knowledge Discovery in Databases,
  Berlin, Germany, September 18-22, 2006, Proceedings}, pages 446--453, 2006.

\bibitem[ABK{\etalchar{+}}06b]{DBLP:conf/kdd/AchtertBKKZ06}
Elke Achtert, Christian B{\"{o}}hm, Hans{-}Peter Kriegel, Peer Kr{\"{o}}ger,
  and Arthur Zimek.
\newblock Deriving quantitative models for correlation clusters.
\newblock In Tina Eliassi{-}Rad, Lyle~H. Ungar, Mark Craven, and Dimitrios
  Gunopulos, editors, {\em Proceedings of the Twelfth {ACM} {SIGKDD}
  International Conference on Knowledge Discovery and Data Mining,
  Philadelphia, PA, USA, August 20-23, 2006}, pages 4--13. {ACM}, 2006.

\bibitem[ABK{\etalchar{+}}07]{DBLP:conf/ssdbm/AchtertBKKZ07}
Elke Achtert, Christian B{\"{o}}hm, Hans{-}Peter Kriegel, Peer Kr{\"{o}}ger,
  and Arthur Zimek.
\newblock On exploring complex relationships of correlation clusters.
\newblock In {\em 19th International Conference on Scientific and Statistical
  Database Management, {SSDBM} 2007, 9-11 July 2007, Banff, Canada,
  Proceedings}, page~7, 2007.

\bibitem[ABKZ06]{DBLP:conf/ssdbm/AchtertBKZ06}
Elke Achtert, Christian B{\"{o}}hm, Peer Kr{\"{o}}ger, and Arthur Zimek.
\newblock Mining hierarchies of correlation clusters.
\newblock In {\em 18th International Conference on Scientific and Statistical
  Database Management, {SSDBM} 2006, 3-5 July 2006, Vienna, Austria,
  Proceedings}, pages 119--128, 2006.

\bibitem[AI06]{DBLP:conf/focs/AndoniI06}
Alexandr Andoni and Piotr Indyk.
\newblock Near-optimal hashing algorithms for approximate nearest neighbor in
  high dimensions.
\newblock In {\em FOCS 2006}, pages 459--468, 2006.

\bibitem[APB13]{DBLP:conf/kdd/AltinigneliPB13}
Muzaffer~Can Altinigneli, Claudia Plant, and Christian B{\"{o}}hm.
\newblock Massively parallel expectation maximization using graphics processing
  units.
\newblock In Inderjit~S. Dhillon, Yehuda Koren, Rayid Ghani, Ted~E. Senator,
  Paul Bradley, Rajesh Parekh, Jingrui He, Robert~L. Grossman, and Ramasamy
  Uthurusamy, editors, {\em The 19th {ACM} {SIGKDD} International Conference on
  Knowledge Discovery and Data Mining, {KDD} 2013, Chicago, IL, USA, August
  11-14, 2013}, pages 838--846. {ACM}, 2013.

\bibitem[Bad08]{DBLP:conf/europar/Bader08}
Michael Bader.
\newblock Exploiting the locality properties of peano curves for parallel
  matrix multiplication.
\newblock In {\em Euro-Par Conference}, pages 801--810, 2008.

\bibitem[BBB{\etalchar{+}}04]{DBLP:journals/bioinformatics/BaumgartnerBBMWOLR04}
Christian Baumgartner, Christian B{\"{o}}hm, Daniela Baumgartner, G.~Marini,
  Klaus Weinberger, B.~Olgem{\"{o}}ller, B.~Liebl, and A.~A. Roscher.
\newblock Supervised machine learning techniques for the classification of
  metabolic disorders in newborns.
\newblock {\em Bioinform.}, 20(17):2985--2996, 2004.

\bibitem[BBB05]{DBLP:journals/jbi/BaumgartnerBB05}
Christian Baumgartner, Christian B{\"{o}}hm, and Daniela Baumgartner.
\newblock Modelling of classification rules on metabolic patterns including
  machine learning and expert knowledge.
\newblock {\em J. Biomed. Informatics}, 38(2):89--98, 2005.

\bibitem[BBBK00]{DBLP:conf/cikm/BohmBBK00}
Christian B{\"{o}}hm, Bernhard Braunm{\"{u}}ller, Markus~M. Breunig, and
  Hans{-}Peter Kriegel.
\newblock High performance clustering based on the similarity join.
\newblock In {\em Proceedings of the 2000 {ACM} {CIKM} International Conference
  on Information and Knowledge Management, McLean, VA, USA, November 6-11,
  2000}, pages 298--305. {ACM}, 2000.

\bibitem[BBK{\etalchar{+}}00]{DBLP:conf/dawak/BerchtoldBKKX00}
Stefan Berchtold, Christian B{\"{o}}hm, Daniel~A. Keim, Hans{-}Peter Kriegel,
  and Xiaowei Xu.
\newblock Optimal multidimensional query processing using tree striping.
\newblock In Yahiko Kambayashi, Mukesh~K. Mohania, and A~Min Tjoa, editors,
  {\em Data Warehousing and Knowledge Discovery, Second International
  Conference, DaWaK 2000, London, UK, September 4-6, 2000, Proceedings}, volume
  1874 of {\em Lecture Notes in Computer Science}, pages 244--257. Springer,
  2000.

\bibitem[BBK{\etalchar{+}}01]{DBLP:conf/icdt/BerchtoldBKK01}
Stefan Berchtold, Christian B{\"{o}}hm, Daniel~A. Keim, Florian Krebs, and
  Hans{-}Peter Kriegel.
\newblock On optimizing nearest neighbor queries in high-dimensional data
  spaces.
\newblock In Jan~Van den Bussche and Victor Vianu, editors, {\em Database
  Theory - {ICDT} 2001, 8th International Conference, London, UK, January 4-6,
  2001, Proceedings}, volume 1973 of {\em Lecture Notes in Computer Science},
  pages 435--449. Springer, 2001.

\bibitem[BBKK01]{epsilongridorder}
Christian B{\"{o}}hm, Bernhard Braunm{\"{u}}ller, Florian Krebs, and Hans-Peter
  Kriegel.
\newblock Epsilon grid order: An algorithm for the similarity join on massive
  high-dimensional data.
\newblock In {\em SIGMOD Conf. 2001}, pages 379--388, 2001.

\bibitem[BBKM00]{DBLP:journals/jiis/BohmBKM00}
Christian B{\"{o}}hm, Stefan Berchtold, Hans{-}Peter Kriegel, and Urs Michel.
\newblock Multidimensional index structures in relational databases.
\newblock {\em J. Intell. Inf. Syst.}, 15(1):51--70, 2000.

\bibitem[BBKS00]{DBLP:conf/adl/BohmBKS00}
Christian B{\"{o}}hm, Bernhard Braunm{\"{u}}ller, Hans{-}Peter Kriegel, and
  Matthias Schubert.
\newblock Efficient similarity search in digital libraries.
\newblock In {\em Proceedings of {IEEE} Advances in Digital Libraries 2000
  {(ADL} 2000), Washington, DC, USA, May 22-24, 2000}, pages 193--199. {IEEE}
  Computer Society, 2000.

\bibitem[BEF08]{DBLP:conf/icde/BryanEF08}
Brent Bryan, Frederick Eberhardt, and Christos Faloutsos.
\newblock Compact similarity joins.
\newblock In {\em {ICDE}}, pages 346--355, 2008.

\bibitem[BFO{\etalchar{+}}09]{DBLP:conf/cikm/BohmFOPW09}
Christian B{\"{o}}hm, Frank Fiedler, Annahita Oswald, Claudia Plant, and Bianca
  Wackersreuther.
\newblock Probabilistic skyline queries.
\newblock In David~Wai{-}Lok Cheung, Il{-}Yeol Song, Wesley~W. Chu, Xiaohua Hu,
  and Jimmy~J. Lin, editors, {\em Proceedings of the 18th {ACM} Conference on
  Information and Knowledge Management, {CIKM} 2009, Hong Kong, China, November
  2-6, 2009}, pages 651--660. {ACM}, 2009.

\bibitem[BFP08]{DBLP:conf/sigmod/BohmFP08}
Christian B{\"{o}}hm, Christos Faloutsos, and Claudia Plant.
\newblock Outlier-robust clustering using independent components.
\newblock In Jason~Tsong{-}Li Wang, editor, {\em Proceedings of the {ACM}
  {SIGMOD} International Conference on Management of Data, {SIGMOD} 2008,
  Vancouver, BC, Canada, June 10-12, 2008}, pages 185--198. {ACM}, 2008.

\bibitem[BFPP07]{DBLP:journals/tkdd/BohmFPP07}
Christian B{\"{o}}hm, Christos Faloutsos, Jia{-}Yu Pan, and Claudia Plant.
\newblock {RIC:} parameter-free noise-robust clustering.
\newblock {\em {ACM} Trans. Knowl. Discov. Data}, 1(3):10, 2007.

\bibitem[BGBF05]{DBLP:journals/jdi/BaumgartnerGBF05}
Christian Baumgartner, Kurt Gautsch, Christian B{\"{o}}hm, and Stephan Felber.
\newblock Functional cluster analysis of {CT} perfusion maps: {A} new tool for
  diagnosis of acute stroke?
\newblock {\em J. Digital Imaging}, 18(3):219--226, 2005.

\bibitem[BGK{\etalchar{+}}07]{DBLP:conf/icde/BohmGKPS07}
Christian B{\"{o}}hm, Michael Gruber, Peter Kunath, Alexey Pryakhin, and
  Matthias Schubert.
\newblock Prover: Probabilistic video retrieval using the gauss-tree.
\newblock In Rada Chirkova, Asuman Dogac, M.~Tamer {\"{O}}zsu, and Timos~K.
  Sellis, editors, {\em Proceedings of the 23rd International Conference on
  Data Engineering, {ICDE} 2007, The Marmara Hotel, Istanbul, Turkey, April
  15-20, 2007}, pages 1521--1522. {IEEE} Computer Society, 2007.

\bibitem[BGO{\etalchar{+}}10]{DBLP:conf/pakdd/BohmGOPPW10}
Christian B{\"{o}}hm, Sebastian Goebl, Annahita Oswald, Claudia Plant, Michael
  Plavinski, and Bianca Wackersreuther.
\newblock Integrative parameter-free clustering of data with mixed type
  attributes.
\newblock In Mohammed~Javeed Zaki, Jeffrey~Xu Yu, Balaraman Ravindran, and
  Vikram Pudi, editors, {\em Advances in Knowledge Discovery and Data Mining,
  14th Pacific-Asia Conference, {PAKDD} 2010, Hyderabad, India, June 21-24,
  2010. Proceedings. Part {I}}, volume 6118 of {\em Lecture Notes in Computer
  Science}, pages 38--47. Springer, 2010.

\bibitem[BHMP09]{DBLP:conf/kdd/BohmHMP09}
Christian B{\"{o}}hm, Katrin Haegler, Nikola~S. M{\"{u}}ller, and Claudia
  Plant.
\newblock Coco: coding cost for parameter-free outlier detection.
\newblock In John F.~Elder IV, Fran{\c{c}}oise Fogelman{-}Souli{\'{e}},
  Peter~A. Flach, and Mohammed~Javeed Zaki, editors, {\em Proceedings of the
  15th {ACM} {SIGKDD} International Conference on Knowledge Discovery and Data
  Mining, Paris, France, June 28 - July 1, 2009}, pages 149--158. {ACM}, 2009.

\bibitem[Bia69]{Bially1969SpacefillingCT}
Theodore Bially.
\newblock Space-filling curves: Their generation and their application to
  bandwidth reduction.
\newblock {\em IEEE Trans. Information Theory}, 15:658--664, 1969.

\bibitem[BK00]{DBLP:conf/edbt/BohmK00}
Christian B{\"{o}}hm and Hans{-}Peter Kriegel.
\newblock Dynamically optimizing high-dimensional index structures.
\newblock In Carlo Zaniolo, Peter~C. Lockemann, Marc~H. Scholl, and Torsten
  Grust, editors, {\em Advances in Database Technology - {EDBT} 2000, 7th
  International Conference on Extending Database Technology, Konstanz, Germany,
  March 27-31, 2000, Proceedings}, volume 1777 of {\em Lecture Notes in
  Computer Science}, pages 36--50. Springer, 2000.

\bibitem[BK01]{DBLP:conf/icde/BohmK01}
Christian B{\"{o}}hm and Hans{-}Peter Kriegel.
\newblock A cost model and index architecture for the similarity join.
\newblock In {\em ICDE}, pages 411--420, 2001.

\bibitem[BK02]{DBLP:conf/icdm/BohmK02}
Christian B{\"{o}}hm and Florian Krebs.
\newblock High performance data mining using the nearest neighbor join.
\newblock In {\em Proceedings of the 2002 {IEEE} International Conference on
  Data Mining {(ICDM} 2002), 9-12 December 2002, Maebashi City, Japan}, pages
  43--50. {IEEE} Computer Society, 2002.

\bibitem[BK03]{DBLP:conf/dexa/BohmK03}
Christian B{\"{o}}hm and Florian Krebs.
\newblock Supporting {KDD} applications by the k-nearest neighbor join.
\newblock In Vladim{\'{\i}}r Mar{\'{\i}}k, Werner Retschitzegger, and Olga
  Step{\'{a}}nkov{\'{a}}, editors, {\em Database and Expert Systems
  Applications, 14th International Conference, {DEXA} 2003, Prague, Czech
  Republic, September 1-5, 2003, Proceedings}, volume 2736 of {\em Lecture
  Notes in Computer Science}, pages 504--516. Springer, 2003.

\bibitem[BK04]{DBLP:journals/kais/BohmK04}
Christian B{\"{o}}hm and Florian Krebs.
\newblock The \emph{k}-nearest neighbour join: Turbo charging the {KDD}
  process.
\newblock {\em Knowl. Inf. Syst.}, 6(6):728--749, 2004.

\bibitem[BKS93]{DBLP:conf/sigmod/BrinkhoffKS93}
Thomas Brinkhoff, Hans{-}Peter Kriegel, and Bernhard Seeger.
\newblock Efficient processing of spatial joins using r-trees.
\newblock In {\em {SIGMOD} Conf. 1993}, pages 237--246, 1993.

\bibitem[BM06]{DBLP:conf/para/BaderM06}
Michael Bader and Christian~E. Mayer.
\newblock Cache oblivious matrix operations using peano curves.
\newblock In {\em PARA Workshop}, pages 521--530, 2006.

\bibitem[BNP{\etalchar{+}}09]{DBLP:journals/tlsdkcs/BohmNPWZ09}
Christian B{\"{o}}hm, Robert Noll, Claudia Plant, Bianca Wackersreuther, and
  Andrew Zherdin.
\newblock Data mining using graphics processing units.
\newblock {\em Trans. Large Scale Data Knowl. Centered Syst.}, 1:63--90, 2009.

\bibitem[BNPZ09]{DBLP:conf/btw/BohmNPZ09}
Christian B{\"{o}}hm, Robert Noll, Claudia Plant, and Andrew Zherdin.
\newblock Indexsupported similarity join on graphics processors.
\newblock In Johann~Christoph Freytag, Thomas Ruf, Wolfgang Lehner, and
  Gottfried Vossen, editors, {\em Datenbanksysteme in Business, Technologie und
  Web {(BTW} 2009), 13. Fachtagung des GI-Fachbereichs "Datenbanken und
  Informationssysteme" (DBIS), Proceedings, 2.-6. M{\"{a}}rz 2009,
  M{\"{u}}nster, Germany}, volume {P-144} of {\em {LNI}}, pages 57--66. {GI},
  2009.

\bibitem[BOPY07]{DBLP:conf/icde/BohmOPY07}
Christian B{\"{o}}hm, Beng~Chin Ooi, Claudia Plant, and Ying Yan.
\newblock Efficiently processing continuous k-nn queries on data streams.
\newblock In Rada Chirkova, Asuman Dogac, M.~Tamer {\"{O}}zsu, and Timos~K.
  Sellis, editors, {\em Proceedings of the 23rd International Conference on
  Data Engineering, {ICDE} 2007, The Marmara Hotel, Istanbul, Turkey, April
  15-20, 2007}, pages 156--165. {IEEE} Computer Society, 2007.

\bibitem[BP08]{DBLP:conf/edbt/BohmP08}
Christian B{\"{o}}hm and Claudia Plant.
\newblock {HISSCLU:} a hierarchical density-based method for semi-supervised
  clustering.
\newblock In Alfons Kemper, Patrick Valduriez, Noureddine Mouaddib, Jens
  Teubner, Mokrane Bouzeghoub, Volker Markl, Laurent Amsaleg, and Ioana
  Manolescu, editors, {\em {EDBT} 2008, 11th International Conference on
  Extending Database Technology, Nantes, France, March 25-29, 2008,
  Proceedings}, volume 261 of {\em {ACM} International Conference Proceeding
  Series}, pages 440--451. {ACM}, 2008.

\bibitem[BPP18]{loopsjournal}
Christian B{\"{o}}hm, Martin Perdacher, and Claudia Plant.
\newblock A novel hilbert curve for cache-locality preserving loops.
\newblock {\em IEEE Transactions on Big Data}, 2018.

\bibitem[BPS06]{DBLP:conf/ssdbm/BohmPS06}
Christian B{\"{o}}hm, Alexey Pryakhin, and Matthias Schubert.
\newblock Probabilistic ranking queries on gaussians.
\newblock In {\em 18th International Conference on Scientific and Statistical
  Database Management, {SSDBM} 2006, 3-5 July 2006, Vienna, Austria,
  Proceedings}, pages 169--178. {IEEE} Computer Society, 2006.

\bibitem[CGL{\etalchar{+}}17]{DBLP:journals/tkde/ChenGLJC17}
Lu~Chen, Yunjun Gao, Xinhan Li, Christian~S. Jensen, and Gang Chen.
\newblock Efficient metric indexing for similarity search and similarity joins.
\newblock {\em {IEEE} Trans. Knowl. Data Eng.}, 29(3):556--571, 2017.

\bibitem[CPZ97]{DBLP:conf/vldb/CiacciaPZ97}
Paolo Ciaccia, Marco Patella, and Pavel Zezula.
\newblock M-tree: An efficient access method for similarity search in metric
  spaces.
\newblock In {\em VLDB'97}, pages 426--435, 1997.

\bibitem[DCHD90]{DBLP:journals/toms/DongarraCHD90}
Jack Dongarra, Jeremy~Du Croz, Sven Hammarling, and Iain~S. Duff.
\newblock A set of level 3 basic linear algebra subprograms.
\newblock {\em {ACM} Trans. Math. Softw.}, 16(1):1--17, 1990.

\bibitem[DEW{\etalchar{+}}12]{DBLP:conf/miccai/DyrbaEWKPOMPBFFHKHKT12}
Martin Dyrba, Michael Ewers, Martin Wegrzyn, Ingo Kilimann, Claudia Plant,
  Annahita Oswald, Thomas Meindl, Michela Pievani, Arun L.~W. Bokde, Andreas
  Fellgiebel, Massimo Filippi, Harald Hampel, Stefan Kl{\"{o}}ppel, Karlheinz
  Hauenstein, Thomas Kirste, and Stefan~J. Teipel.
\newblock Combining {DTI} and {MRI} for the automated detection of alzheimer's
  disease using a large european multicenter dataset.
\newblock In Pew{-}Thian Yap, Tianming Liu, Dinggang Shen, Carl{-}Fredrik
  Westin, and Li~Shen, editors, {\em Multimodal Brain Image Analysis - Second
  International Workshop, {MBIA} 2012, Held in Conjunction with {MICCAI} 2012,
  Nice, France, October 1-5, 2012. Proceedings}, volume 7509 of {\em Lecture
  Notes in Computer Science}, pages 18--28. Springer, 2012.

\bibitem[DGSZ03]{DBLP:journals/mta/DohnalGSZ03}
Vlastislav Dohnal, Claudio Gennaro, Pasquale Savino, and Pavel Zezula.
\newblock D-index: Distance searching index for metric data sets.
\newblock {\em Multimedia Tools Appl.}, 21(1):9--33, 2003.

\bibitem[DGZ03]{DBLP:conf/dexa/DohnalGZ03}
Vlastislav Dohnal, Claudio Gennaro, and Pavel Zezula.
\newblock Similarity join in metric spaces using ed-index.
\newblock In {\em {DEXA} 2003}, pages 484--493, 2003.

\bibitem[DS01]{DBLP:conf/kdd/DittrichS01}
Jens{-}Peter Dittrich and Bernhard Seeger.
\newblock {GESS:} a scalable similarity-join algorithm for mining large data
  sets in high dimensional spaces.
\newblock In {\em {SIGKDD}}, pages 47--56, 2001.

\bibitem[DTL18]{DBLP:conf/sigmod/DengT018}
Dong Deng, Yufei Tao, and Guoliang Li.
\newblock Overlap set similarity joins with theoretical guarantees.
\newblock In {\em {SIGMOD} Conf. 2018}, pages 905--920, 2018.

\bibitem[FAB{\etalchar{+}}18]{DBLP:journals/pvldb/FierABLF18}
Fabian Fier, Nikolaus Augsten, Panagiotis Bouros, Ulf Leser, and
  Johann{-}Christoph Freytag.
\newblock Set similarity joins on mapreduce: An experimental survey.
\newblock {\em {PVLDB}}, 11(10):1110--1122, 2018.

\bibitem[FHK{\etalchar{+}}12]{DBLP:conf/kdd/FengHKBP12}
Jing Feng, Xiao He, Bettina Konte, Christian B{\"{o}}hm, and Claudia Plant.
\newblock Summarization-based mining bipartite graphs.
\newblock In Qiang Yang, Deepak Agarwal, and Jian Pei, editors, {\em The 18th
  {ACM} {SIGKDD} International Conference on Knowledge Discovery and Data
  Mining, {KDD} '12, Beijing, China, August 12-16, 2012}, pages 1249--1257.
  {ACM}, 2012.

\bibitem[FLPR99]{DBLP:conf/focs/FrigoLPR99}
Matteo Frigo, Charles~E. Leiserson, Harald Prokop, and Sridhar Ramachandran.
\newblock Cache-oblivious algorithms.
\newblock In {\em FOCS 1999}, pages 285--298, 1999.

\bibitem[FRR14]{DBLP:journals/bmcbi/FerreiraRR14}
Miguel Ferreira, Nuno Roma, and Lu{\'{\i}}s M.~S. Russo.
\newblock Cache-oblivious parallel {SIMD} viterbi decoding for sequence search
  in {HMMER}.
\newblock {\em {BMC} Bioinformatics}, 15:165, 2014.

\bibitem[GHPB14]{DBLP:conf/icdm/GoeblHPB14}
Sebastian Goebl, Xiao He, Claudia Plant, and Christian B{\"{o}}hm.
\newblock Finding the optimal subspace for clustering.
\newblock In Ravi Kumar, Hannu Toivonen, Jian Pei, Joshua~Zhexue Huang, and
  Xindong Wu, editors, {\em 2014 {IEEE} International Conference on Data
  Mining, {ICDM} 2014, Shenzhen, China, December 14-17, 2014}, pages 130--139.
  {IEEE} Computer Society, 2014.

\bibitem[HLLY07]{DBLP:conf/sigmod/HeLLY07}
Bingsheng He, Yinan Li, Qiong Luo, and Dongqing Yang.
\newblock Easedb: a cache-oblivious in-memory query processor.
\newblock In {\em {SIGMOD} Conf. 2007}, pages 1064--1066, 2007.

\bibitem[HOK18]{DBLP:journals/tjs/HoOK18}
ThienLuan Ho, Seungrohk Oh, and Hyunjin Kim.
\newblock New algorithms for fixed-length approximate string matching and
  approximate circular string matching under the hamming distance.
\newblock {\em The Journal of Supercomputing}, 74(5):1815--1834, 2018.

\bibitem[Kal13]{DBLP:journals/vldb/Kalashnikov13}
Dmitri~V. Kalashnikov.
\newblock Super-ego: fast multi-dimensional similarity join.
\newblock {\em {VLDB} J.}, 22(4):561--585, 2013.

\bibitem[KP03]{DBLP:conf/dasfaa/KalashnikovP03}
Dmitri~V. Kalashnikov and Sunil Prabhakar.
\newblock Similarity join for low-and high-dimensional data.
\newblock In {\em {(DASFAA} '03)}, pages 7--16, 2003.

\bibitem[KP07]{DBLP:journals/is/KalashnikovP07}
Dmitri~V. Kalashnikov and Sunil Prabhakar.
\newblock Fast similarity join for multi-dimensional data.
\newblock {\em Inf. Syst.}, 32(1):160--177, 2007.

\bibitem[KS00]{DBLP:journals/tkde/KoudasS00}
Nick Koudas and Kenneth~C. Sevcik.
\newblock High dimensional similarity joins: Algorithms and performance
  evaluation.
\newblock {\em {IEEE} Trans. Knowl. Data Eng.}, 12(1):3--18, 2000.

\bibitem[LSS08]{DBLP:conf/icde/LiebermanSS08}
Michael~D. Lieberman, Jagan Sankaranarayanan, and Hanan Samet.
\newblock A fast similarity join algorithm using graphics processing units.
\newblock In {\em {ICDE}}, pages 1111--1120, 2008.

\bibitem[LWU16]{DBLP:conf/waim/LiWU16}
Ye~Li, Jian Wang, and Leong~Hou U.
\newblock Multidimensional similarity join using mapreduce.
\newblock In {\em Web-Age Information Management}, pages 457--468, 2016.

\bibitem[MBK{\etalchar{+}}10]{DBLP:journals/tog/MoonBKCKBNY10}
Bochang Moon, Yongyoung Byun, Tae{-}Joon Kim, Pio Claudio, Hye{-}Sun Kim,
  Yun{-}Ji Ban, Seung~Woo Nam, and Sung{-}Eui Yoon.
\newblock Cache-oblivious ray reordering.
\newblock {\em {ACM} Trans. Graph.}, 29(3), 2010.

\bibitem[MGP12]{DBLP:conf/icdm/MaiGP12}
Son~T. Mai, Sebastian Goebl, and Claudia Plant.
\newblock A similarity model and segmentation algorithm for white matter fiber
  tracts.
\newblock In Mohammed~Javeed Zaki, Arno Siebes, Jeffrey~Xu Yu, Bart Goethals,
  Geoffrey~I. Webb, and Xindong Wu, editors, {\em 12th {IEEE} International
  Conference on Data Mining, {ICDM} 2012, Brussels, Belgium, December 10-13,
  2012}, pages 1014--1019. {IEEE} Computer Society, 2012.

\bibitem[MHF{\etalchar{+}}15]{DBLP:journals/kais/MaiHFPB15}
Son~T. Mai, Xiao He, Jing Feng, Claudia Plant, and Christian B{\"{o}}hm.
\newblock Anytime density-based clustering of complex data.
\newblock {\em Knowl. Inf. Syst.}, 45(2):319--355, 2015.

\bibitem[MJZ17]{DBLP:journals/concurrency/MaJZ17}
Youzhong Ma, Shijie Jia, and Yongxin Zhang.
\newblock A novel approach for high-dimensional vector similarity join query.
\newblock {\em Concurrency and Computation: Practice and Experience}, 29(5),
  2017.

\bibitem[MS18]{DBLP:conf/sigmod/McCauley018}
Samuel McCauley and Francesco Silvestri.
\newblock Adaptive mapreduce similarity joins.
\newblock In {\em {SIGMOD} Workshop on Algorithms and Systems for MapReduce and
  Beyond}, pages 4:1--4:4, 2018.

\bibitem[PB11]{DBLP:conf/kdd/PlantB11}
Claudia Plant and Christian B{\"{o}}hm.
\newblock {INCONCO:} interpretable clustering of numerical and categorical
  objects.
\newblock In Chid Apt{\'{e}}, Joydeep Ghosh, and Padhraic Smyth, editors, {\em
  Proceedings of the 17th {ACM} {SIGKDD} International Conference on Knowledge
  Discovery and Data Mining, San Diego, CA, USA, August 21-24, 2011}, pages
  1127--1135. {ACM}, 2011.

\bibitem[PBTB06]{DBLP:journals/bioinformatics/PlantBTB06}
Claudia Plant, Christian B{\"{o}}hm, Bernhard Tilg, and Christian Baumgartner.
\newblock Enhancing instance-based classification with local density: a new
  algorithm for classifying unbalanced biomedical data.
\newblock {\em Bioinform.}, 22(8):981--988, 2006.

\bibitem[Pla12]{DBLP:conf/kdd/Plant12}
Claudia Plant.
\newblock Dependency clustering across measurement scales.
\newblock In Qiang Yang, Deepak Agarwal, and Jian Pei, editors, {\em The 18th
  {ACM} {SIGKDD} International Conference on Knowledge Discovery and Data
  Mining, {KDD} '12, Beijing, China, August 12-16, 2012}, pages 361--369.
  {ACM}, 2012.

\bibitem[PM18]{DBLP:journals/corr/abs-1803-04765}
Nicolas Papernot and Patrick~D. McDaniel.
\newblock Deep k-nearest neighbors: Towards confident, interpretable and robust
  deep learning.
\newblock {\em CoRR}, abs/1803.04765, 2018.

\bibitem[PPSS17]{DBLP:journals/algorithmica/PaghPSS17}
Rasmus Pagh, Ninh Pham, Francesco Silvestri, and Morten St{\"{o}}ckel.
\newblock I/o-efficient similarity join.
\newblock {\em Algorithmica}, 78(4):1263--1283, 2017.

\bibitem[PR09]{DBLP:journals/jda/ParedesR09}
Rodrigo Paredes and Nora Reyes.
\newblock Solving similarity joins and range queries in metric spaces with the
  list of twin clusters.
\newblock {\em J. Discrete Algorithms}, 7(1):18--35, 2009.

\bibitem[Pru86]{Prusinkiewicz:1986:GAL:16564.16608}
P~Prusinkiewicz.
\newblock Graphical applications of l-systems.
\newblock In {\em Proceedings on Graphics Interface '86/Vision Interface '86},
  pages 247--253, 1986.

\bibitem[PS14]{DBLP:conf/sisap/PearsonS14}
Spencer~S. Pearson and Yasin~N. Silva.
\newblock Index-based {R-S} similarity joins.
\newblock In {\em {SISAP}}, pages 106--112, 2014.

\bibitem[PWZ09]{DBLP:conf/icdm/PlantWZ09}
Claudia Plant, Afra~M. Wohlschl{\"{a}}ger, and Andrew Zherdin.
\newblock Interaction-based clustering of multivariate time series.
\newblock In Wei Wang, Hillol Kargupta, Sanjay Ranka, Philip~S. Yu, and Xindong
  Wu, editors, {\em {ICDM} 2009, The Ninth {IEEE} International Conference on
  Data Mining, Miami, Florida, USA, 6-9 December 2009}, pages 914--919. {IEEE}
  Computer Society, 2009.

\bibitem[SD89]{DBLP:conf/sigmod/SchneiderD89}
Donovan~A. Schneider and David~J. DeWitt.
\newblock A performance evaluation of four parallel join algorithms in a
  shared-nothing multiprocessor environment.
\newblock In {\em {SIGMOD} Conf. 1989}, pages 110--121, 1989.

\bibitem[SHB{\etalchar{+}}13]{DBLP:journals/tkde/ShaoHBYP13}
Junming Shao, Xiao He, Christian B{\"{o}}hm, Qinli Yang, and Claudia Plant.
\newblock Synchronization-inspired partitioning and hierarchical clustering.
\newblock {\em {IEEE} Trans. Knowl. Data Eng.}, 25(4):893--905, 2013.

\bibitem[SLLF17]{DBLP:conf/icde/ShangLLF17}
Zeyuan Shang, Yaxiao Liu, Guoliang Li, and Jianhua Feng.
\newblock K-join: Knowledge-aware similarity join.
\newblock In {\em {ICDE}}, pages 23--24, 2017.

\bibitem[SMY{\etalchar{+}}12]{SHAO20122756}
Junming Shao, Nicholas Myers, Qinli Yang, Jing Feng, Claudia Plant, Christian
  Böhm, Hans Förstl, Alexander Kurz, Claus Zimmer, Chun Meng, Valentin Riedl,
  Afra Wohlschläger, and Christian Sorg.
\newblock Prediction of alzheimer's disease using individual structural
  connectivity networks.
\newblock {\em Neurobiology of Aging}, 33(12):2756--2765, 2012.

\bibitem[SPYB11]{DBLP:conf/icdm/ShaoPYB11}
Junming Shao, Claudia Plant, Qinli Yang, and Christian B{\"{o}}hm.
\newblock Detection of arbitrarily oriented synchronized clusters in
  high-dimensional data.
\newblock In Diane~J. Cook, Jian Pei, Wei Wang, Osmar~R. Za{\"{\i}}ane, and
  Xindong Wu, editors, {\em 11th {IEEE} International Conference on Data
  Mining, {ICDM} 2011, Vancouver, BC, Canada, December 11-14, 2011}, pages
  607--616. {IEEE} Computer Society, 2011.

\bibitem[SS83]{10.1007/BFb0000120}
Rani Siromoney and K.~G. Subramanian.
\newblock Space-filling curves and infinite graphs.
\newblock In Hartmut Ehrig, Manfred Nagl, and Grzegorz Rozenberg, editors, {\em
  Graph-Grammars and Their Application to Computer Science}, pages 380--391,
  Berlin, Heidelberg, 1983. Springer Berlin Heidelberg.

\bibitem[SWY{\etalchar{+}}17]{DBLP:journals/kais/ShaoWYPB17}
Junming Shao, Xinzuo Wang, Qinli Yang, Claudia Plant, and Christian B{\"{o}}hm.
\newblock Synchronization-based scalable subspace clustering of
  high-dimensional data.
\newblock {\em Knowl. Inf. Syst.}, 52(1):83--111, 2017.

\bibitem[WMP13]{DBLP:conf/kdd/WangMP13}
Ye~Wang, Ahmed Metwally, and Srinivasan Parthasarathy.
\newblock Scalable all-pairs similarity search in metric spaces.
\newblock In {\em {SIGKDD}}, pages 829--837, 2013.

\bibitem[XWL08]{DBLP:journals/pvldb/XiaoWL08}
Chuan Xiao, Wei Wang, and Xuemin Lin.
\newblock Ed-join: an efficient algorithm for similarity joins with edit
  distance constraints.
\newblock {\em {PVLDB}}, 1(1):933--944, 2008.

\bibitem[XWL{\etalchar{+}}11]{DBLP:journals/tods/XiaoWLYW11}
Chuan Xiao, Wei Wang, Xuemin Lin, Jeffrey~Xu Yu, and Guoren Wang.
\newblock Efficient similarity joins for near-duplicate detection.
\newblock {\em {ACM} Trans. Database Syst.}, 36(3):15:1--15:41, 2011.

\bibitem[YGPB16]{DBLP:conf/kdd/YeGPB16}
Wei Ye, Sebastian Goebl, Claudia Plant, and Christian B{\"{o}}hm.
\newblock {FUSE:} full spectral clustering.
\newblock In Balaji Krishnapuram, Mohak Shah, Alexander~J. Smola, Charu~C.
  Aggarwal, Dou Shen, and Rajeev Rastogi, editors, {\em Proceedings of the 22nd
  {ACM} {SIGKDD} International Conference on Knowledge Discovery and Data
  Mining, San Francisco, CA, USA, August 13-17, 2016}, pages 1985--1994. {ACM},
  2016.

\bibitem[YMHP16]{DBLP:conf/icdm/YeMHP16}
Wei Ye, Samuel Maurus, Nina Hubig, and Claudia Plant.
\newblock Generalized independent subspace clustering.
\newblock In Francesco Bonchi, Josep Domingo{-}Ferrer, Ricardo Baeza{-}Yates,
  Zhi{-}Hua Zhou, and Xindong Wu, editors, {\em {IEEE} 16th International
  Conference on Data Mining, {ICDM} 2016, December 12-15, 2016, Barcelona,
  Spain}, pages 569--578. {IEEE} Computer Society, 2016.

\bibitem[YNL{\etalchar{+}}17]{DBLP:journals/tkde/YuNLWY17}
Chenyun Yu, Sarana Nutanong, Hangyu Li, Cong Wang, and Xingliang Yuan.
\newblock A generic method for accelerating lsh-based similarity join
  processing.
\newblock {\em {IEEE} Trans. Knowl. Data Eng.}, 29(4):712--726, 2017.

\bibitem[YSS{\etalchar{+}}12]{DBLP:journals/envsoft/YangSSBP12}
Qinli Yang, Junming Shao, Miklas Scholz, Christian B{\"{o}}hm, and Claudia
  Plant.
\newblock Multi-label classification models for sustainable flood retention
  basins.
\newblock {\em Environ. Model. Softw.}, 32:27--36, 2012.

\bibitem[ZRDW16]{DBLP:conf/sigmod/ZhaoRDW16}
Weijie Zhao, Florin Rusu, Bin Dong, and Kesheng Wu.
\newblock Similarity join over array data.
\newblock In {\em {SIGMOD} Conf. 2016}, pages 2007--2022, 2016.

\end{thebibliography}
\end{document}